\documentclass{llncs}
\usepackage[utf8]{inputenc}
\usepackage{graphicx}
\usepackage{url}
\usepackage{amsmath}
\usepackage{xcolor}
\usepackage{overpic}
\usepackage{subfigure}
\usepackage{todonotes}
\usepackage{xspace}
\usepackage{float}
\usepackage{amssymb} 
\usepackage{subfigure} 
\usepackage{epstopdf}

\newcommand\nao{NAO}

\newcommand\ie{\mbox{i.\,e.}\xspace}
\newcommand\eg{\mbox{e.\,g.}\xspace}
\newcommand\secref[1]{\mbox{Sect.}~\ref{#1}}
\newcommand\figref[1]{\mbox{Fig.}~\ref{#1}}
\newcommand\tabref[1]{\mbox{Tab.}~\ref{#1}}

\newcommand{\Figref}[1]{Fig. \ref{#1}}
\newcommand{\Tabref}[1]{Tab. \ref{#1}}
\newcommand{\equref}[1]{eq. \ref{#1}}

\usepackage{cite}
\usepackage{microtype} 
\title{Kick Motions for the NAO Robot using Dynamic Movement Primitives}
\author{Arne Böckmann \and Tim Laue}
\institute{Universität Bremen, Fachbereich 3 -- Mathematik und Informatik, \\
Postfach 330 440, 28334 Bremen, Germany \\
E-Mail: {\ttfamily $\{$arneboe,tlaue$\}$@informatik.uni-bremen.de}}

\begin{document}

\maketitle

\begin{abstract}
In this paper, we present the probably first application of the popular \emph{Dynamic Movement Primitives (DMP)} approach to the domain of soccer-playing humanoid robots. DMPs are known for their ability to imitate previously demonstrated motions as well as to flexibly adapt to unforeseen changes to the desired trajectory with respect to speed and direction. As demonstrated in this paper, this makes them a useful approach for describing kick motions. Furthermore, we present a mathematical motor model that compensates for the NAO robot's motor control delay as well as a novel minor extension to the DMP formulation. The motor model is used in the calculation of the Zero Moment Point (ZMP), which is needed to keep the robot in balance while kicking. All approaches have been evaluated on real NAO robots.
\end{abstract}

\section{Introduction}

Kick motions are an essential part of robot soccer. In recent years, the speed of the game has increased a lot with most teams now being able to stably walk at high speeds. Thus, fights for the ball are more common. A flexible kick motion that is able to adapt to different and changing ball locations as well as to different kick speeds on the fly while keeping the robot in balance during pushes from other robots is a huge advantage in such situations.

Several methods to design and execute flexible kick motions have already been developed. For instance, Müller et al. \cite{BIKE-RoboCup-2010} model the kick foot trajectory using hand-crafted piecewise Bézier curves, which are modified on the fly to adapt to different ball positions. However, handcrafting Bézier curves is a complex and time consuming task. Wenk et al. \cite{RC-Wenk-Roefer-14} tackle this problem by automatically inferring trajectories based on the ball position, kick velocity, and kick direction. While this method works, it does not allow the user to influence the resulting trajectory, \ie creating special purpose kicks like backward kicks is not possible.

In this paper, we present a middle ground between the two above-mentioned approaches: A kick motion that can be hand-crafted easily by using kinematic teach-in or be created by a multitude of optimization algorithms while retaining the ability to adapt to different ball positions and kick velocities.
This is done by using a modified version of Dynamic Movement Primitives (DMPs)\cite{ijspeert2013dynamical} to describe the kick trajectory.
During the kick, the robot is dynamically balanced using a Linear Quadratic Regulator (LQR) with previews to keep the Zero Moment Point (ZMP) inside the support polygon. This involves a new way of estimating the ZMP based on a model of the motor behavior of the NAO.

The remainder of the paper is organized as follows: Section \ref{sec:motor} introduces the motor model that is the basis of the ZMP calculation, \secref{sec:balance} explains the ZMP estimation and introduces the balancing algorithm while \secref{sec:dmp} introduces DMPs and explains how we use them to model a kick trajectory. Sections \ref{sec:eval} and \ref{sec:conclusion} wrap up the paper with an evaluation of the kick motion and a conclusion.

\section{Model-Based Motor Position Prediction}
\label{sec:motor}
As described below, the NAO's motor response delay is usually 30 ms. Thus, the ZMP balancer needs to take into account that the motor will not be at the currently measured position when the current command reaches the motor. Our solution to this problem is to predict the current motor position using a mathematical model and to use this prediction in our control algorithms.

\subsection{Determining the NAO's Motor Response Delay}
\label{subsec:delay}
The \nao's motors are position-controlled using the proprietary NaoQi software. It processes the commands and relays them to an ARM-7 micro controller at 100 Hz. The controller distributes the commands over RS-485 to several dsPIC micro controllers which are responsible for controlling the actual motors. Measured motor positions travel back the same chain\cite{naoDesign}. This chain together with the slow control rate of 100 Hz induces a delay between sending a command and being able to measure a reaction of the motor.

\begin{figure}[t]
  \centering
  \includegraphics[width=\textwidth]{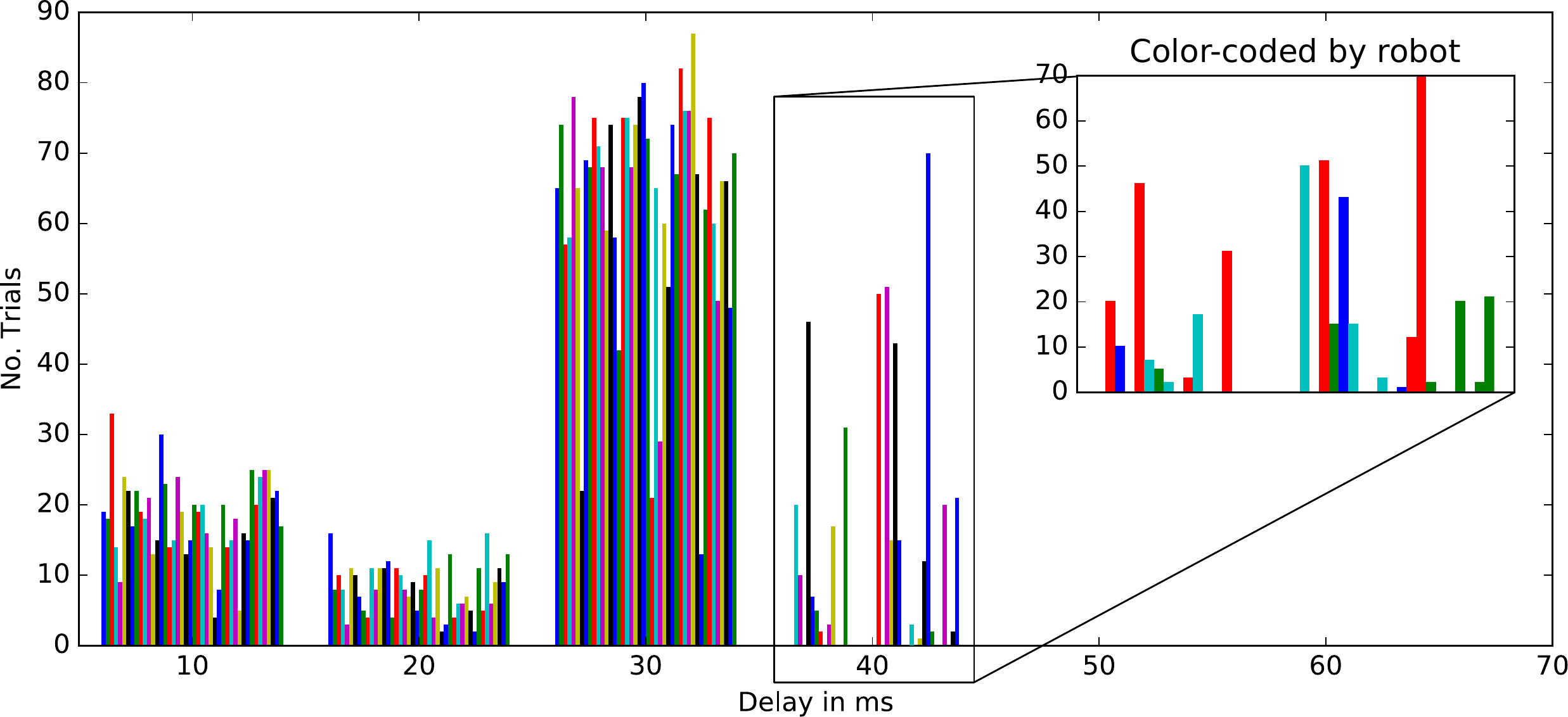}
  \caption{Histograms of motor response delay. Each bar is one motor of one robot. The zoomed part shows the motors that responded after 40ms, color-coded by robot. The experiment was conducted using the leg motors of four robots and repeating each trial 100 times. Thus, 4400 responses were measured in total.}
    \label{fig:motor_delay_histogram}
\end{figure}

To determine the actual delay, a motor is moved from a resting position into a random direction and the time between sending the command and measuring a movement is recorded. A movement is registered as soon as the measured motor position deviates from the position that the motor was in when the command was issued. No threshold is used. For measuring, the internal sensor is used. 
 This is done 100 times for each leg motor of four different NAOs, thus, we get 4400 measurements in total.
As shown in \figref{fig:motor_delay_histogram}, the vast majority of motor reactions occurs at 30 milliseconds.
The measured average distance of the reactions at 10 and 20 ms is $0.087^\circ$. This is below the maximum accuracy of the motors, which is $0.1^\circ$. Therefore, we can assume that the measurements at 10 and 20 ms are due to sensor noise. However, the measurements at 40 ms cannot be discarded as noise.
Taking a closer look, it seems that some joints in some robots are more prone to responding after 40 ms than others, suggesting that hardware wear or defects might cause a delayed measurement.

Thus, for a fully repaired robot it is safe to assume that the motor response delay is 30 ms. Actually, the delay might be anywhere between 20 and 30 ms, but due to the 100 Hz duty cycle, more precise measurements are not possible.

\subsection{A Model to Estimate the Motor Position}
We propose to model the behavior of a motor as second order dynamical system based on a mass spring damper:
\begin{align}
  T^2\ddot{y}(t) + 2DT min(\dot{y}(t), V_{max}) + y(t) = u(t), \text{~~} T, D, y, u \in \mathbb{R}
\end{align}
$T$ is the time constant, $y(t)$ is the motor position at time $t$, $D$ is a dampening constant, $V_{max}$ is the maximum motor velocity that is used to limit $\dot{y}(t)$, and $u(t)$ is the requested motor position at time $t$.
 
The parameters $(T, D, V_{max})$ need to be set in a way that the model optimally mimics a motor. This can be achieved by minimizing the error function $J$:
\begin{align}
  J = \sum_{s \in S} \sum_{i=0}^{|s|} d(i) (m(i) - s(i))^2
\end{align}
$S$ is a set of step responses for a given motor, $|s|$ the number of measurements in step response $s$, $m(i)$ the position of the model at the $i$-th step, $s(i)$ the actual motor position at the $i$-th step and $d(i) = 0.85^i$ a decay function.

The decay function $d$ emphasizes the short term model quality over the long term, \ie we prefer parameters that provide a better short term prediction over parameters that provide an overall good prediction. This is done because in our use case the model is only used to predict a short amount of time. 

\begin{figure}[t]
    \centering
    \subfigure[]{\includegraphics[width=0.48\textwidth]{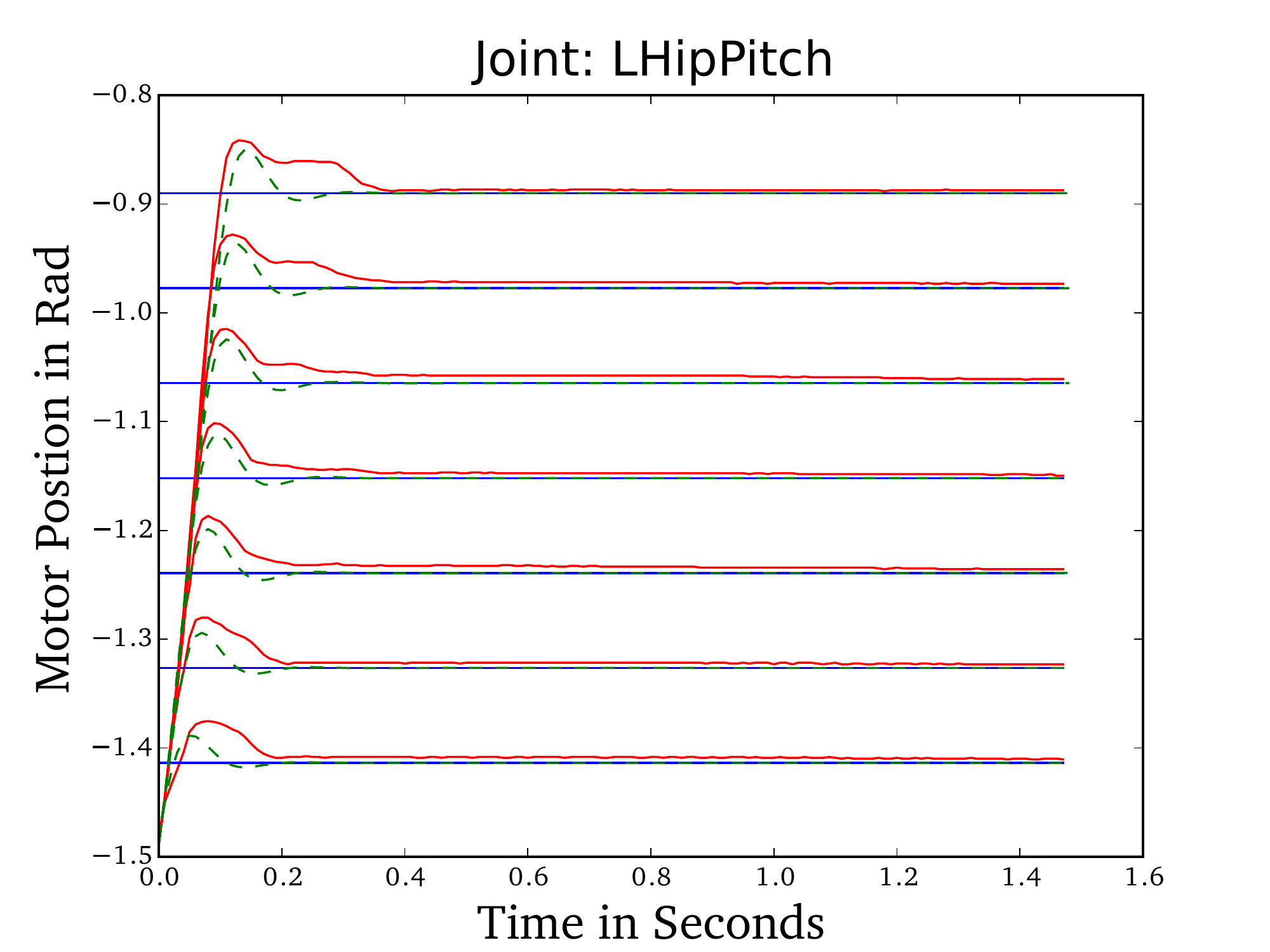} \label{fig:step_response_fit}}
    \subfigure[]{\includegraphics[width=0.48\textwidth]{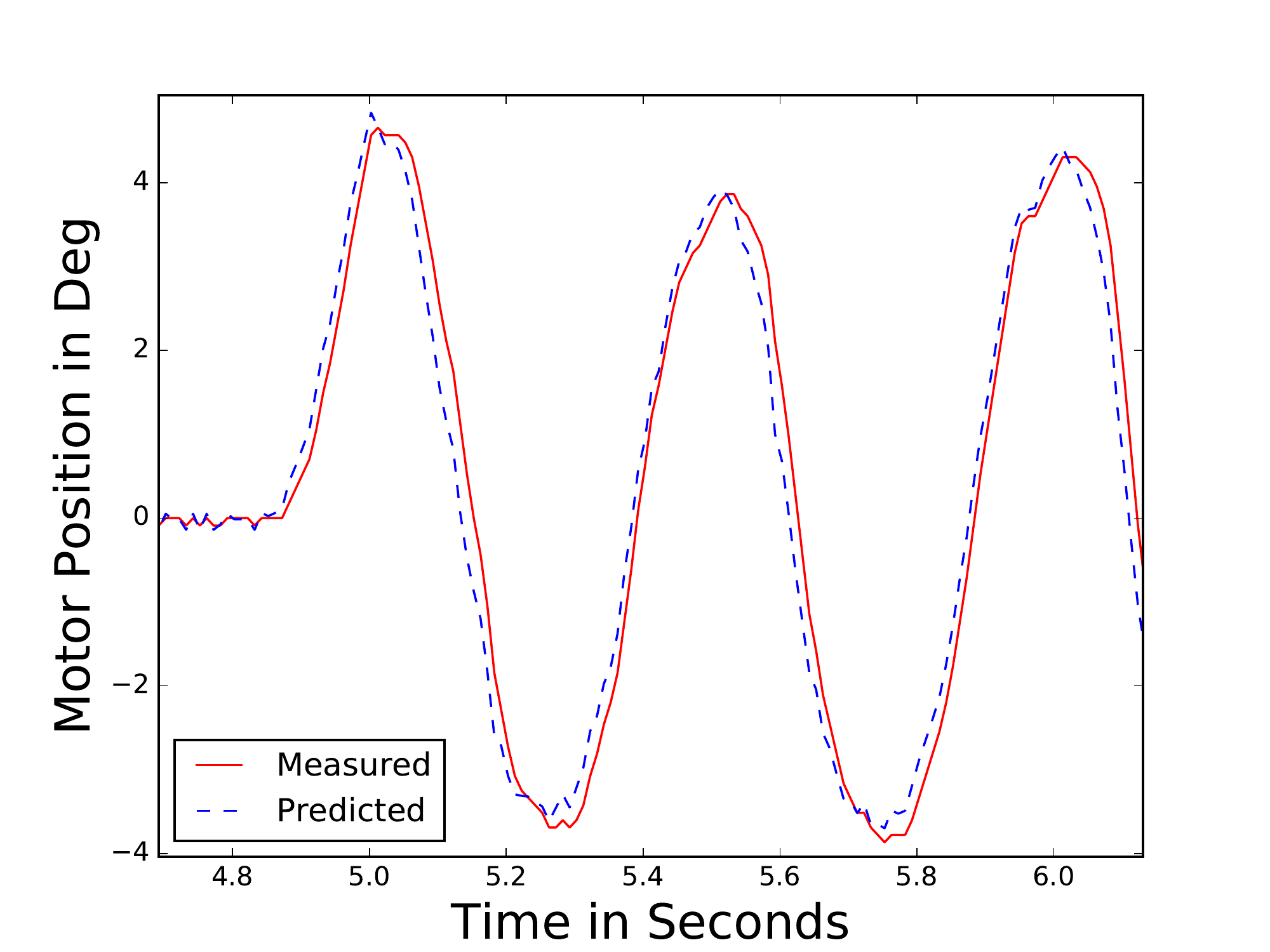} \label{fig:step_response_walk_shift}}

    \caption{(a) Step responses and their optimal fit. The blue lines are the step functions, red are the step responses, and green the response of the best fitting model. The model of the depicted LHipPitch joint has an error that is slightly above average (see \Tabref{tab:model_walk_error}). (b) Prediction and actual motor response of a motor while walking}
\end{figure}

Sets of step responses can be generated by applying step functions, which jump from zero to their respective values instantly, with different step heights to the motor. \Figref{fig:step_response_fit} shows a set of step responses that has been recorded and the respective optimal model response. For fitting, the first three samples of the step response should be ignored to make up for the motor response delay.

\subsection{Model Evaluation}
To show that the model can be used to predict the real world motor positions of the \nao, we compared the model response and the actual motor position while executing a five second walking motion. For the comparison, the real motor values have been shifted in time to remove the measurement delay. \Figref{fig:step_response_walk_shift} shows an excerpt of the experiment of the LHipRoll motor.
The average absolute error over all leg joints is $0.185^{\circ}$ with a variance of $0.0573$. Thus, the model seems to be able to predict the motor behavior with sufficient accuracy. Detailed results for each motor can be seen in \tabref{tab:model_walk_error}.

\begin{table}[t]
\parbox{.49\linewidth}{
\centering
    \begin{tabular}{|c|c|c|}
    \hline
    Joint       & Avg. error & Variance\\\hline\hline
    LAnklePitch                   & $0.196^{\circ}$        & 0.04\\\hline
    LAnkleRoll                    & $0.153^{\circ}$        & 0.02\\\hline
    LHipPitch                     & $0.216^{\circ}$        & 0.053\\\hline
    LHipRoll                      & $0.111^{\circ}$        & 0.009\\\hline
    LHipYawPitch                  & $0.048^{\circ}$        & 0.003\\\hline
    LKneePitch                    & $0.363^{\circ}$        & 0.145\\\hline
    RAnklePitch                   & $0.216^{\circ}$        & 0.054\\\hline
    RAnkleRoll                    & $0.170^{\circ}$        & 0.025\\\hline
    RHipPitch                     & $0.307^{\circ}$        & 0.092\\\hline
    RHipRoll                      & $0.097^{\circ}$        & 0.007\\\hline
    RKneePitch                    & $0.297^{\circ}$        & 0.118\\\hline
    \end{tabular}
    
    \vspace{1mm} 
\caption{Error in leg motor predictions}
\label{tab:model_walk_error}
}
\parbox{.49\linewidth}{
\centering
    \begin{tabular}{|c|c|c|}
    \hline
    Robot       &  \begin{tabular}[b]{@{}c@{}} Avg. error\\ over all leg joints\end{tabular} & Variance\\\hline\hline
    Original                   & $0.185^{\circ}$        & 0.057\\\hline
    NAO 1                    & $0.159^{\circ}$        & 0.055\\\hline
    NAO 2                     & $0.183^{\circ}$        & 0.065\\\hline
    NAO 3                      & $0.179^{\circ}$        & 0.056\\\hline
    NAO 4                  & $0.181^{\circ}$        & 0.071\\\hline
    NAO 5                    & $0.197^{\circ}$        & 0.061\\\hline
    \end{tabular}
    \vspace{1mm}
\caption{Errors when applying model parameters that have been optimized for one NAO to five different NAOs}
\label{tab:model_walk_error_portability}
}
\end{table}

We were also interested in the portability of the model parameters. As documented in \tabref{tab:model_walk_error_portability}, a repetition of the experiment on different robots (always using the same model again) was successful.

\section{ZMP-based Balancing}
\label{sec:balance}
The robot needs to be kept in balance while kicking. 
A good measure for the balance of a robot is the Zero Moment Point (ZMP). A robot is said to be dynamically stable if the ZMP is inside the support polygon\cite{vukobratovic2004zero}.

The center of pressure of the support polygon and the Zero Moment Point are coincident\cite{sardain2004forces}. Therefore, it is possible to use the pressure sensors under the NAO's feet to measure the ZMP, if it exists. However, the sensors are quite inaccurate and often faulty. To avoid using these sensors, we estimate the ZMP $(z_x, z_y)^T$ based on the cart-table model proposed by Kajita et al.\cite{kajitaCartTable}:
\begin{align} 
 \begin{pmatrix} z_x\\z_y\end{pmatrix} = 
 \begin{pmatrix} c_x\\c_y\end{pmatrix} -  \frac{c_z}{g} \begin{pmatrix}\ddot{c_x}\\ \ddot{c_y}\end{pmatrix} 
\end{align}
$(c_x, c_y, c_z)^T$ is the center of mass (COM) and $g\approx9.81$ the gravitational force.
Due to the measurement delay and the high sensor noise, we estimate the COM using the motor model and forward kinematics $(c_x^m, c_y^m, c_z^m)^T$. The motor model is initialized using sensor readings at the beginning of the motion. While the kick motion is being executed, the model is not updated from sensor readings.

\begin{figure}[tb]
    \centering
    \subfigure[]{\includegraphics[width=0.49\textwidth]{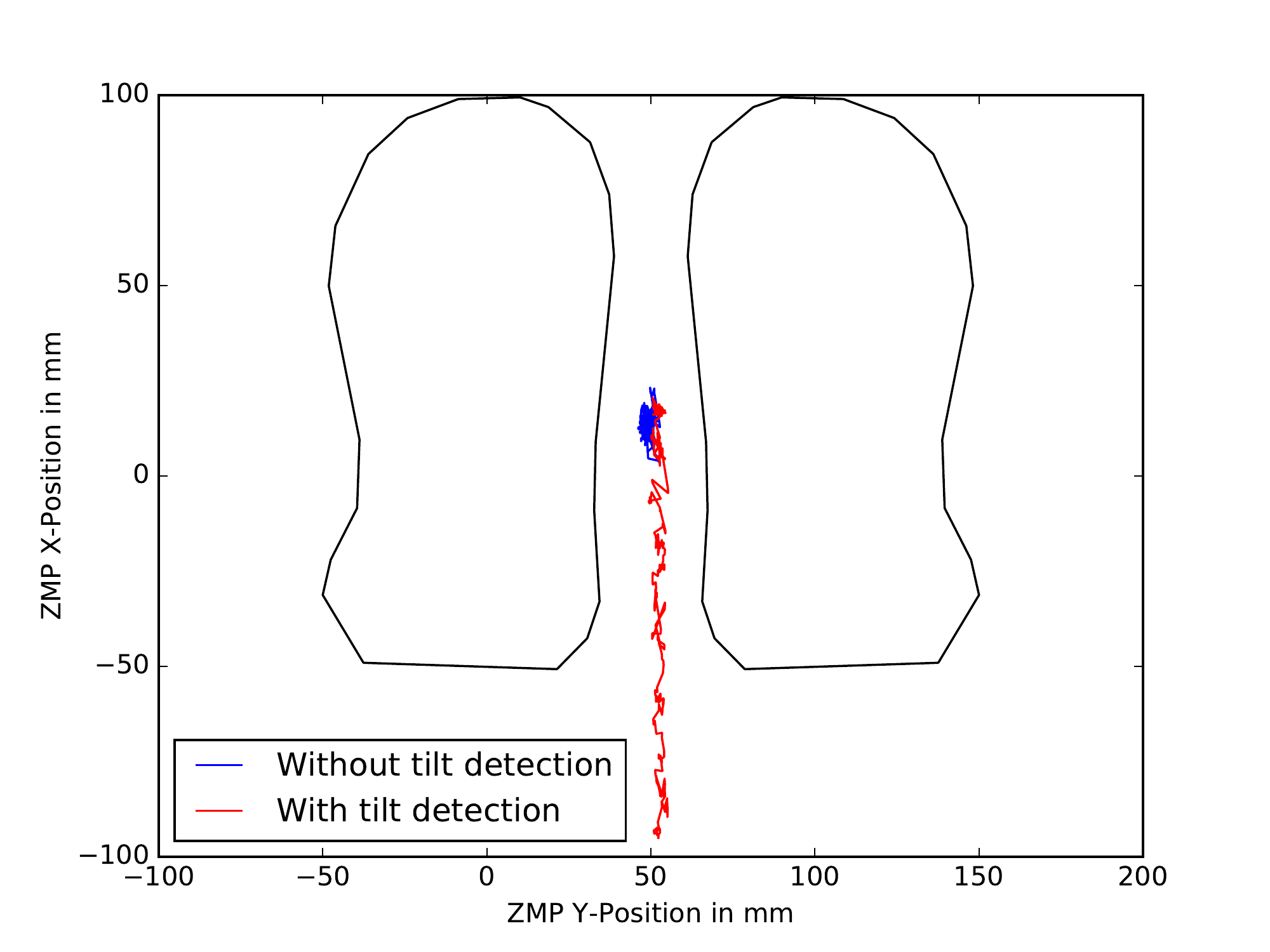}\label{fig:zmp_shift:a}}
    \subfigure[]{\includegraphics[width=0.49\textwidth]{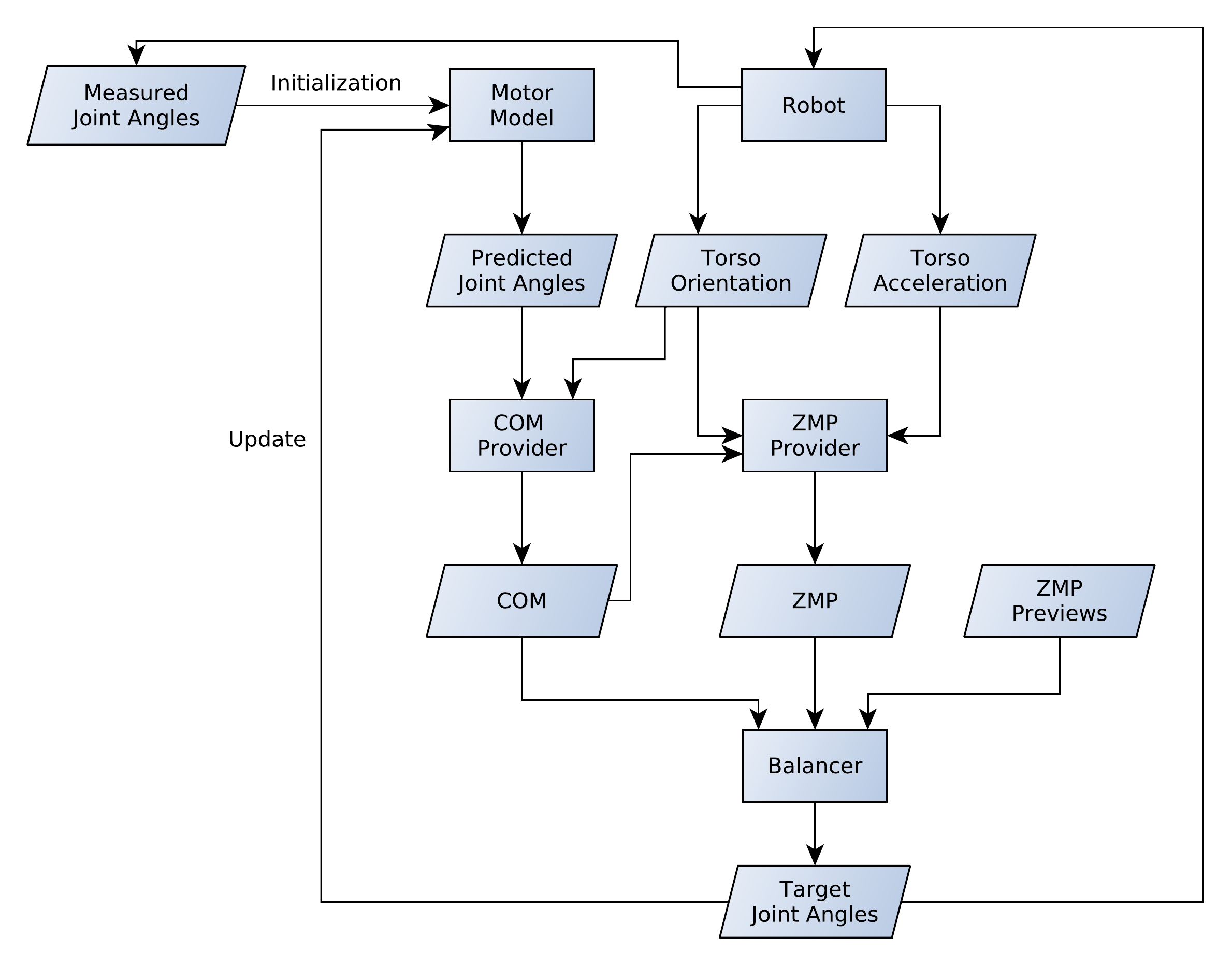}\label{fig:balance_overview}}

    \caption{(a) The red and blue lines show the estimated ZMP position with and without tilt detection while the robot is being tilted backwards. (b) The balancing process.}\label{fig:zmp_shift}
\end{figure}

Tilting the robot over the edges of the supporting foot does not influence the estimated ZMP. To detect such situations, we calculate the scaled difference $P\Delta\Theta=P(\gamma - \phi)$ between the expected torso orientation $\gamma$ as provided by the motor model and the measured torso orientation $\phi$ as provided by the IMU and scale the COM accordingly\cite{alcaraz2013robust}. The unitless constant factor $P$ needs to be adjusted manually. We chose $P=(30, -30)^T$. \Figref{fig:zmp_shift:a} shows how the ZMP behaves with and without tilt detection.

To be able to measure outside influences, \eg someone pushing the robot, we replaced the COM acceleration by the acceleration of the torso $\ddot{o}$ as measured by the NAO's IMU. This can be done because the COM is usually inside the torso and thus both accelerations are similar. 

Thus, the final ZMP is calculated by:
\begin{align} 
 \begin{pmatrix} z_x\\z_y\end{pmatrix} = 
 P\Delta\Theta\begin{pmatrix} c_x^m\\c_y^m\end{pmatrix} -  \frac{c_z^m}{g} \begin{pmatrix}\ddot{o_x}\\ \ddot{o_y}\end{pmatrix} 
\end{align}

To finally balance the robot, an LQR preview controller as described in \cite{tasse2010, tasse2013, RC-Wenk-Roefer-14} has been implemented. 
 The inputs of the controller are the current ZMP and current COM as well as the next 50 desired ZMP positions. \Figref{fig:balance_overview} shows an overview of the balancing sub-system.

\section{Describing Kicks Using Dynamic Movement Primitives}
\label{sec:dmp}
The kick trajectory is described by using Dynamic Movement Primitives (DMPs)\cite{schaal2006dynamic, ijspeert2013dynamical}.
For the sake of simplicity, this chapter only considers one-dimensional DMPs but they can be easily scaled to $n$ dimensions as long as the dimensions are independent of each other, \ie for translational movements. For rotational movements, a special DMP formulation has been introduced by Ude et al.\cite{dmpUde}. However, for kick motions it is sufficient to simply keep the foot level to the ground, thus no rotational DMP was used in this paper.

DMPs model goal-directed movements as weakly non-linear dynamical systems. They consist of the canonical system and the transformation system.

The \textit{canonical system} $s$ describes the phase of the movement:
\begin{align}
      \tau\dot{s} &= -\alpha_ss \label{eq:canonical_system}
\end{align}
The phase $s$ replaces the time in the transformation system. Intuitively, it drives the transformation system similar to a clock\cite{Muelling30012013}.
$s$ conventionally starts at 1 and monotonically converges to zero. $\tau$ is the execution time of the movement. As long as $\tau$ remains constant, a closed solution for the canonical system exists\cite{Muelling30012013}:
\begin{align}
  s(t) = exp(\frac{-\alpha_s}{\tau} t) \label{eq:canonicalSystemDirect}
\end{align}
$\alpha_s$ determines how fast $s$ converges. The value of $\alpha_s$ has to be chosen in a way that $s$ is sufficiently close to zero at the end of the execution. We chose $\alpha_s$ by setting $s(\tau) = 0.01$ and solving \eqref{eq:canonicalSystemDirect} for $\alpha_s$.

The \textit{transformation system} is defined by two first order differential equations:
\begin{align}
	\tau\dot{z} &= \alpha_z(\beta_z(g-y)-z) + sf(s)\label{eq:transformation_system_1}\\
	\tau\dot{y} &= z\label{eq:transformation_system_2}
\end{align}
$\tau$ is the execution time of the movement, $\alpha_z$ and $\beta_z$ are damping constants, $g$ is the goal position of the movement, $y$ is the current position of the movement, $s$ is the current phase as defined by \equref{eq:canonical_system} and $f(s)$ is called the forcing term. 

The dampening constants are set for critical dampening, \ie $\beta_z = \alpha_z/4$. We chose $\beta_z=25$ and $\alpha_z=6.25$, but the exact values do not matter as long as the system is critically dampened.

The forcing term defines the movement's shape. With $f(s) = 0$, the system is just a PD controller converging to $g$ and reaching it at time $\tau$. One could say that $f$ superimposes its shape onto the PD controller.
To ensure that $f$ does not keep the system from reaching the desired goal position, $f$ is scaled by the phase, thus diminishing the influence of $f$ towards the end of the movement.
To be able to express arbitrary movements, $f$ is typically chosen to be a radial basis function approximator:
\begin{align}
f(s) &= \frac{\sum_{i=1}^N \psi_i(s)w_i}{\sum_{i=1}^N \psi_i(s)}, w_i \in \mathbb{R} \label{eq:rbf}
\end{align}
$\psi_i$ is the i-th Gaussian radial basis function with mean $c_i$ and variance $\sigma_i^2$:
\begin{align}
	\psi_i(s) = exp(-\frac{1}{2\sigma_i^2}(s - c_i)^2)
\end{align}
The weights $w_i$ can be chosen to create any desired function and thus define the shape of the whole movement. Different learning and optimization approaches can be used to find the weights for a certain movement, \eg Schaal et al. \cite{schaal2003control} describe  how to imitate a given trajectory.

Since the goal position $g$ is part of \equref{eq:transformation_system_1}, the shape also depends on $g$. This means that the weights can only force the system into a certain shape for one specific value of $g$. When $g$ changes, the shape ``bends'' to reach the new goal position. In general, this is undesired behavior and is solved by scaling $f$ if $g$ changes.
\begin{figure}[tb]
    \centering
    \subfigure[]{
      \includegraphics[width=0.475\textwidth]{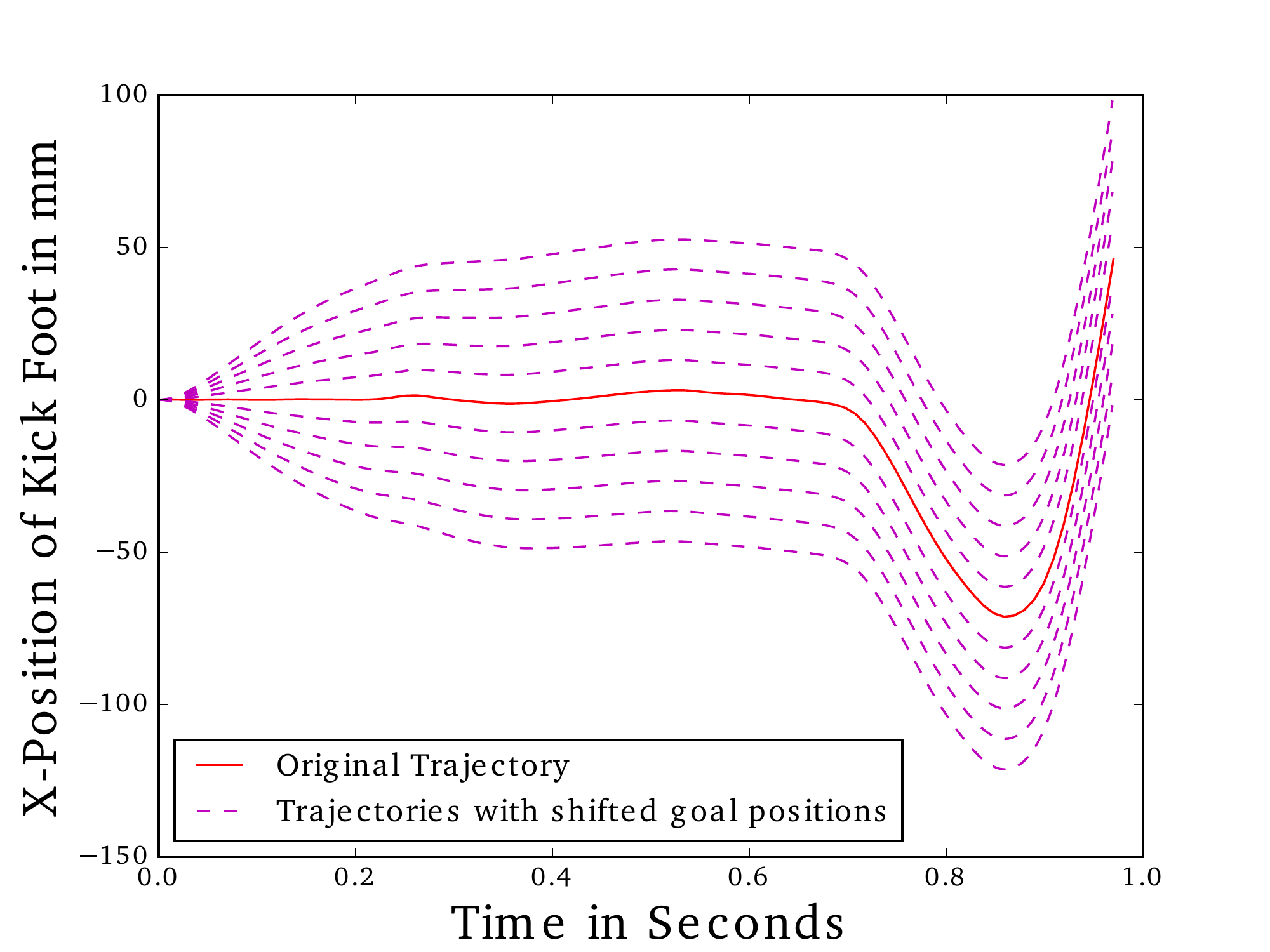}
      \label{fig:shift_goal_position_no_scaling:a}
    }
    \subfigure[]{
      \includegraphics[width=0.475\textwidth]{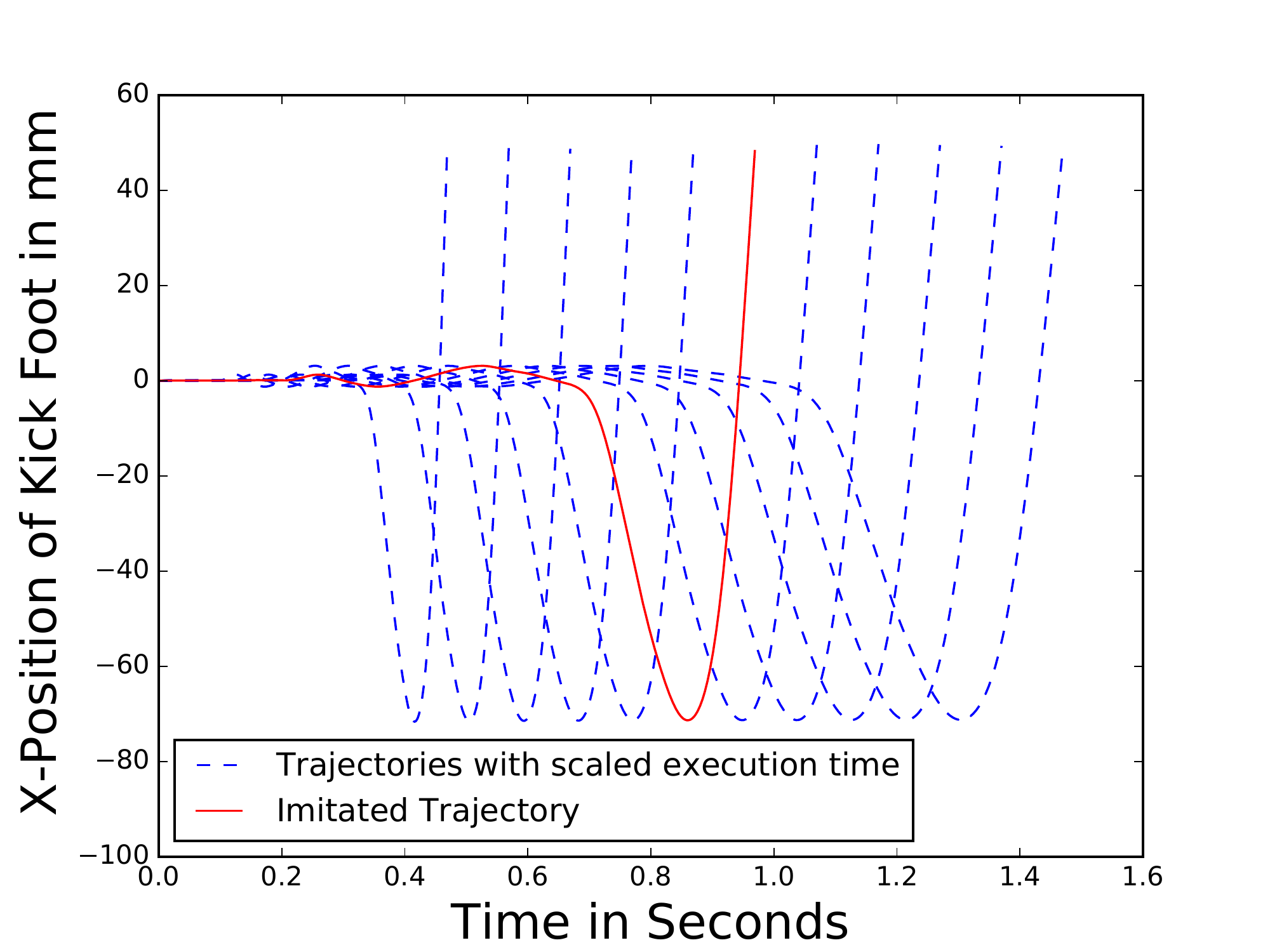}
      \label{fig:shift_goal_position_no_scaling:b}
    }
    \caption{(a) Behavior of the kick motion when the goal position is adapted. (b) Temporal scaling capability of the DMP.}
    \label{fig:shift_goal_position_no_scaling}
\end{figure}
However, we found that kick motions ``bend'' in a natural way (\figref{fig:shift_goal_position_no_scaling}), \ie if the goal position is moved closer to the start position, the robot will swing further back and vice versa. This is exactly the behavior that one would expect from a kicking motion because it ensures that the distance between the inflection point and the goal position remains the same. This is important because the kicking velocity depends on this distance.

Thus, the DMP formulation allows us to define an arbitrary movement in task space and later scale its execution time as well as to move the goal position while retaining a sane shape.

A major downside of the formulation is that the final velocity is always zero making it unsuitable for kick motions because kick motions need to reach the target with a specific velocity. Solutions to this problem were proposed by
Kober et al. \cite{koberDmp} and Mülling et al. \cite{Muelling30012013}.
They replaced the goal $g$ in \equref{eq:transformation_system_1} with the position, velocity and acceleration of a moving target $g_p$. 
\begin{align}
  \tau\dot{z} = \alpha_z(\beta_z(g_p - y) + \tau\dot{g_p} - z) + \tau^2\ddot{g_p} + sf(s) \label{eq:muellingDmp}
\end{align}
While Kober et al. used a target that is moving on a straight line, Mülling et al. used a fifth order polynomial.
\begin{align} 
  g_p(t) = \sum_{i=0}^5b_it,~~~~
  \dot{g_p}(t) = \sum_{i=i}^5ib_it^{i-1},~~~~
  \ddot{g_p}(t) =\sum_{i=2}^5(i^2-i)b_it^{i-2}
\end{align}

The coefficients $b_i$ are calculated by applying the bounding conditions:
 \begin{align}
  g_p(t_0) &= y_0, ~~~ \dot{g_p}(t_0) = \dot{y_0}, ~~~ \ddot{g_p}(t_0) = \ddot{y_0}\\
  g_p(\tau) &= g, ~~~ \dot{g_p}(\tau) = \dot{g}, ~~~ \ddot{g_p}(\tau) = 0
\end{align}
Due to the time dependency, the coefficients need to be recalculated if $\tau$ changes.

\begin{figure}[tb]
    \centering
    \subfigure[]{
    \includegraphics[width=0.475\textwidth]{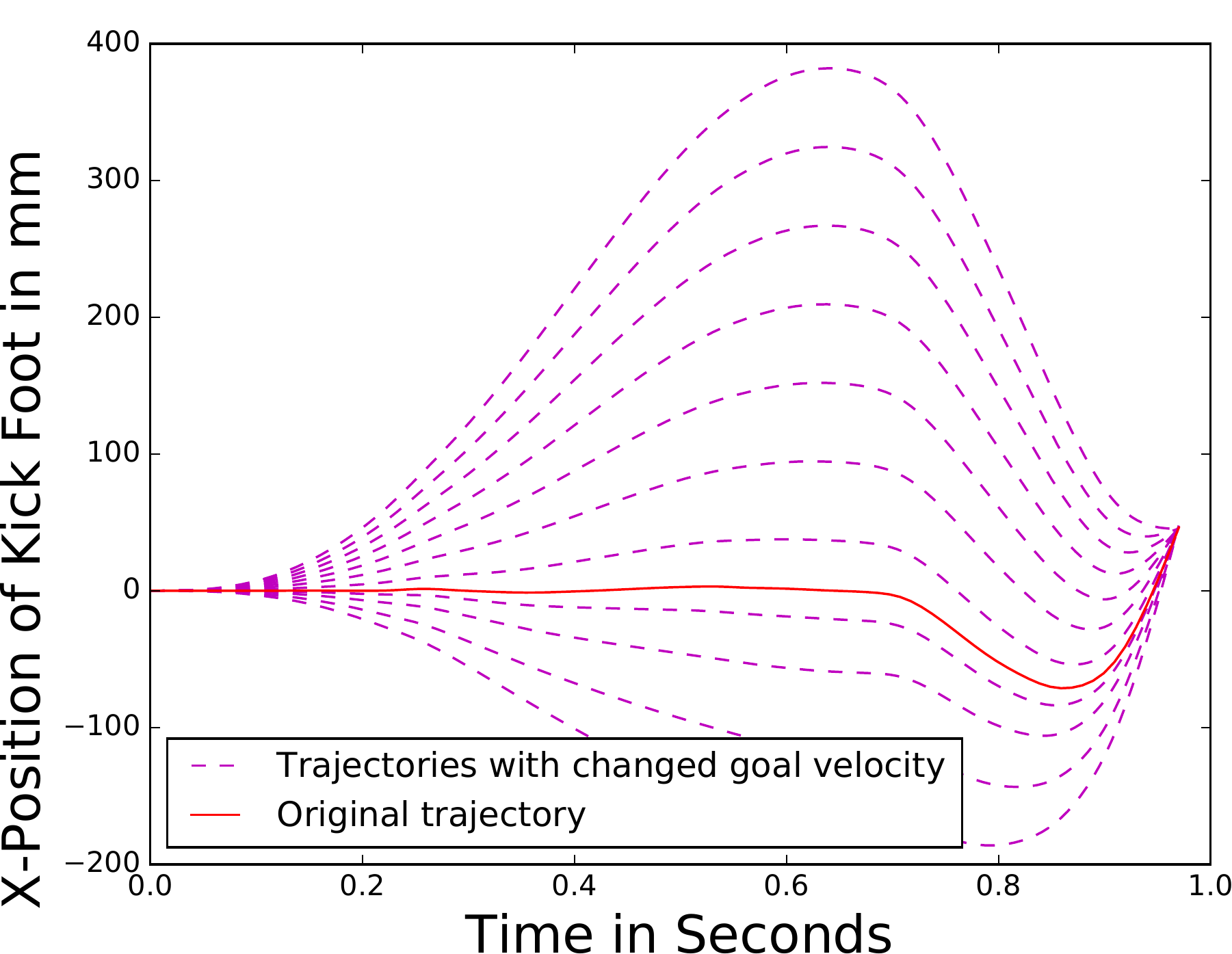}
    \label{fig:muelling_dmp_speed_scale:a}
    }
    \subfigure[]{
    \includegraphics[width=0.475\textwidth]{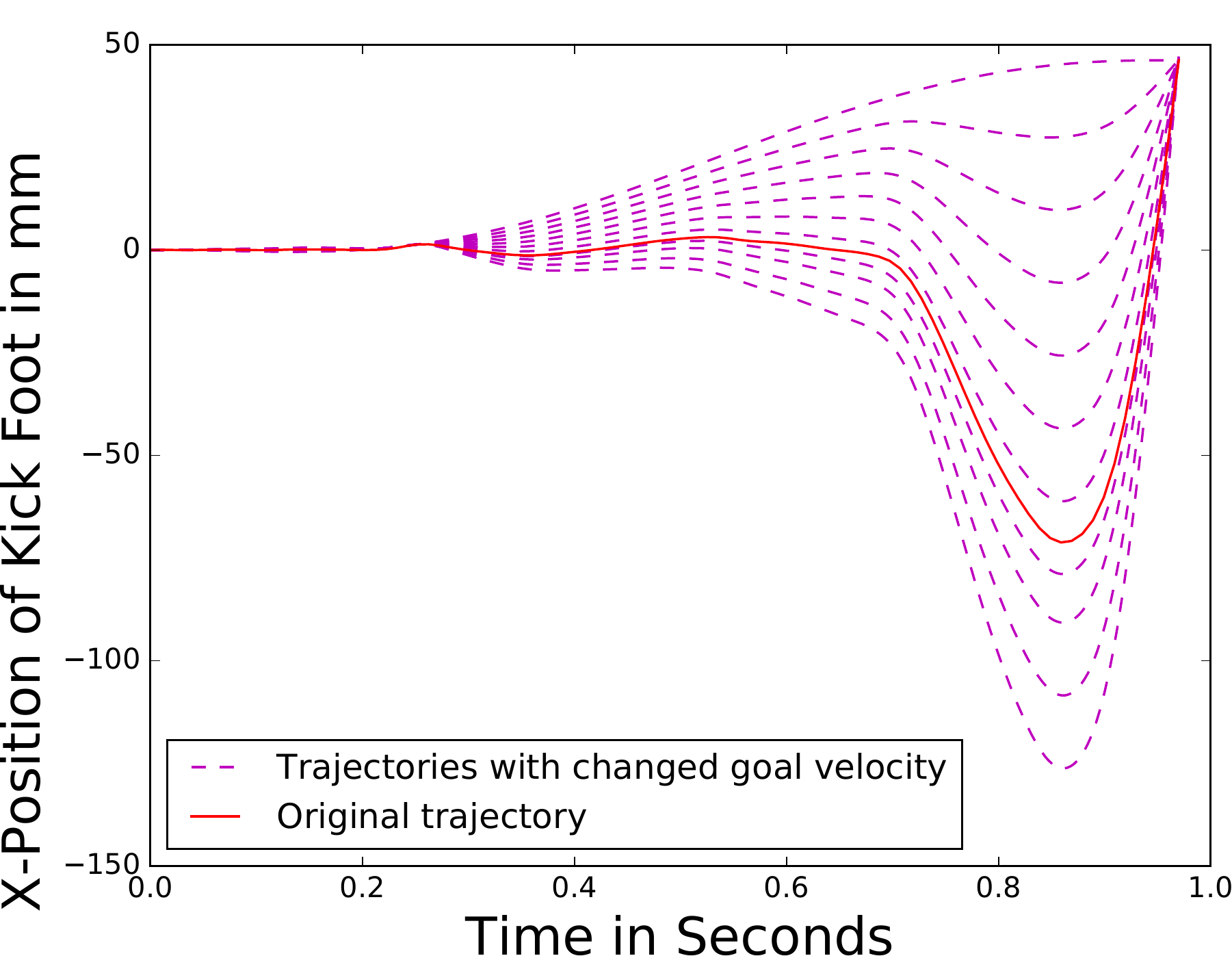}
    \label{fig:muelling_dmp_speed_scale:b}
    }

    \caption{Reaction of the DMP to changes in the final velocity. (a) Shows the reaction of the original DMP while (b) shows how the DMP reacts with our new scaling term $A$.}
    \label{fig:muelling_dmp_speed_scale}
\end{figure}

In this way, a new parameter $\dot{g}$ is introduced. It represents the velocity at the end of the movement. However, the weights now depend on $\dot{g}$ as well. This means that if the goal velocity is changed, the shape of the movement will change. As shown in \figref{fig:muelling_dmp_speed_scale:a}, the trajectory reacts to changes in the goal velocity with huge changes and becomes inexecutable.
We propose to fix this by scaling the forcing term with the novel factor $A =  (\dot{g}_{new} - \dot{y}_0) / (\dot{g} - \dot{y}_0)$, where $\dot{y}_0$ is the starting velocity of the trajectory and $\dot{g}_{new}$ is the new goal velocity.
\Figref{fig:muelling_dmp_speed_scale:b} shows that this produces much better results. If the velocity is increased, the wind up phase gets longer, if it is reduced, the wind up phase gets shorter until it completely disappears if the requested goal velocity is zero. This is exactly the behavior that one would expect from a kick motion.

Thus, the final form of the DMP used in our experiments is:
\begin{align}
  \tau\dot{z} &= \alpha_z(\beta_z(g_p - y) + \tau\dot{g_p} - z) + \tau^2\ddot{g_p} + sf(s)A\\
  \tau\dot{y} &= z\label{eq:dmp_final}\\
  \tau\dot{s} &= -\alpha_ss
\end{align}
This DMP responds well to changes in goal position and goal velocity. It is noteworthy that both parameters can be changed mid-execution without causing discontinuities. The implementation used in our experiments has been released as part of the B-Human Code Release 2015\cite{codeRel2015} and is available online\footnote{{\scriptsize  \url{https://github.com/bhuman/BHumanCodeRelease/tree/master/Src/Tools/Motion}}}.

\section{Evaluation}
\label{sec:eval} 

\begin{table}[tb]
  \begin{center}
    \begin{tabular}{|l|l|l|l|l|}
    \hline
                        & B-Human left   & Imitated left & B-Human right & Imitated right\\\hline\hline
Avg. distance           & 5.3 m          & 4.3 m         & 4.61 m        & 3.6 m\\
Avg. angular deviation  & $5.62^\circ$   & $8.06^\circ$  & $9.69^\circ$  & $7.48^\circ$\\
Avg. ball location (cm) & (-2.53, 496.93)&(-1.63, 430.03)& (7.52, 453.53)&(4.41, 361.56)\\
Royston H-value         &  4.48          & 0.407         & 1.62          & 3.08\\
Royston p-value         &  0.106         & 0.812         & 0.43          & 0.000062\\
Is normal distributed   &  Yes           & Yes           & Yes           & No\\\hline
    \end{tabular}
  \end{center}
  \caption{Kick distance comparision between the B-Human kick motion of 2015 and an imitation of that motion. The data for the imitated kick with the right leg contained two outliers. If those outliers are removed the result is normally distributed as well.}
  \label{tab:imitatedKickResults}
\end{table}

\begin{figure}[tb]
    \centering
    \subfigure[]{
    \includegraphics[width=0.475\textwidth]{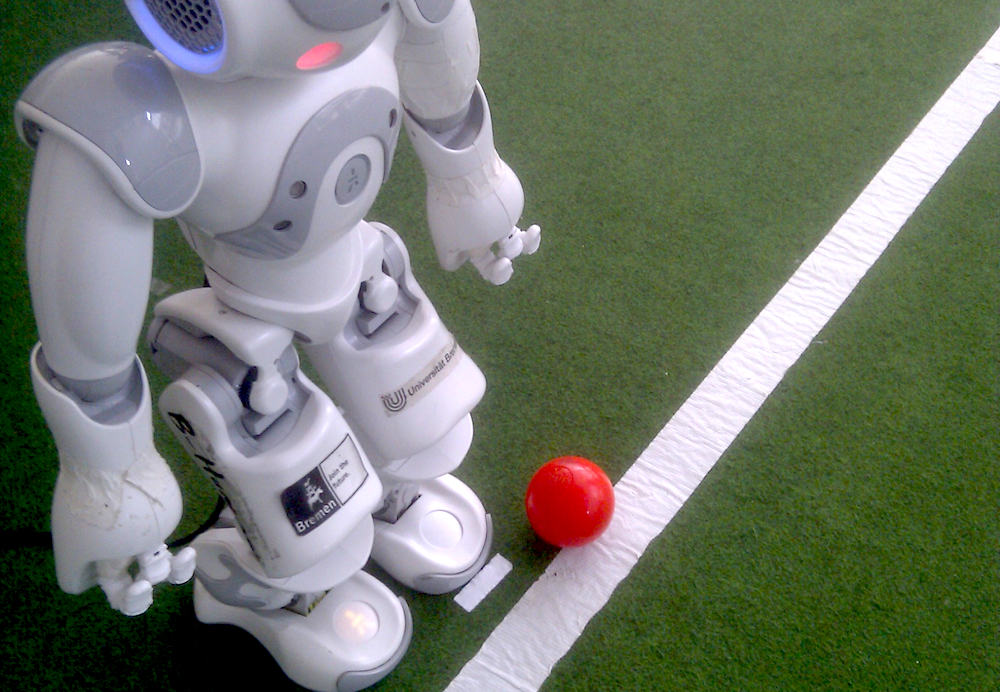}
    \label{fig:kick_experiments:a}
    }
    \subfigure[]{
    \includegraphics[width=0.475\textwidth]{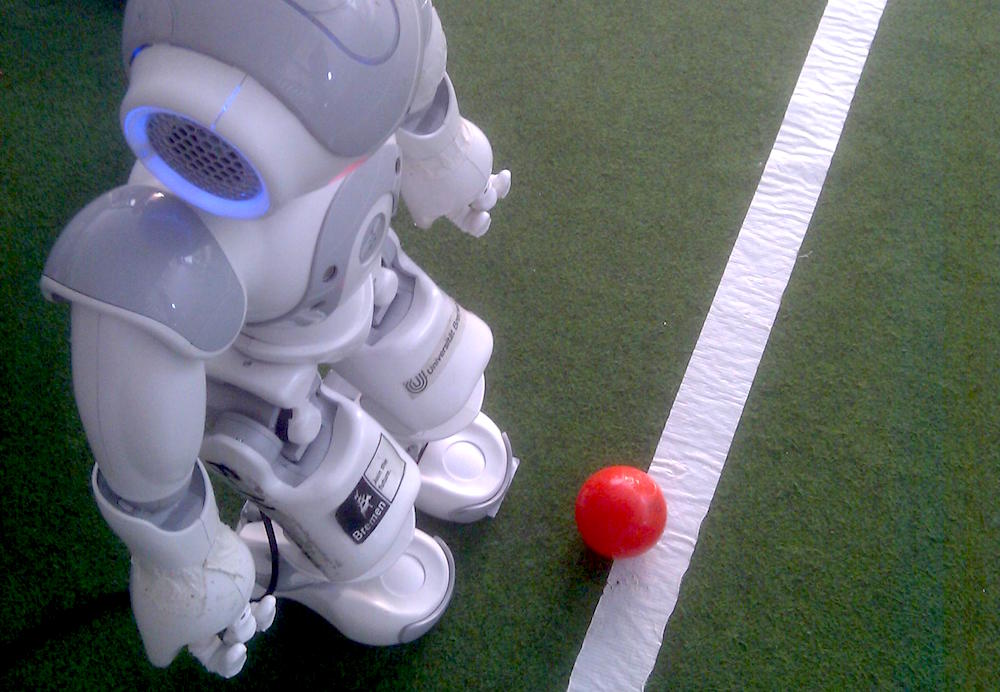}
    \label{fig:kick_experiments:b}
    }

    \caption{Ball position generalization experiments: In (a) the ball was moved 65 mm to the left, in (b) it was moved 80 mm to the front.}
    \label{fig:kick_experiments}
\end{figure}

Several experiments have been done to evaluate the kick motion. The setup is identical for all experiments: The robot is standing at the side line of the field and kicks the ball into the field, as depicted in \figref{fig:kick_experiments}. All experiments have been done with the official RoboCup SPL ball of 2015, a Mylec street hockey ball that is 65 mm in diameter and weighs 55 g. Each experiment consists of 30 kicks.

We used the Royston H-Test\cite{royston1983some} with a significance level of 0.05 to determine that the measured kick distances are normally distributed. Normally distributed kick results indicate that the results have only been influenced by natural noise, \ie there is probably no systematic error in the test setup or the implementation.

\begin{figure}[tb]
    \centering
    \subfigure[]{
    \includegraphics[width=0.475\textwidth]{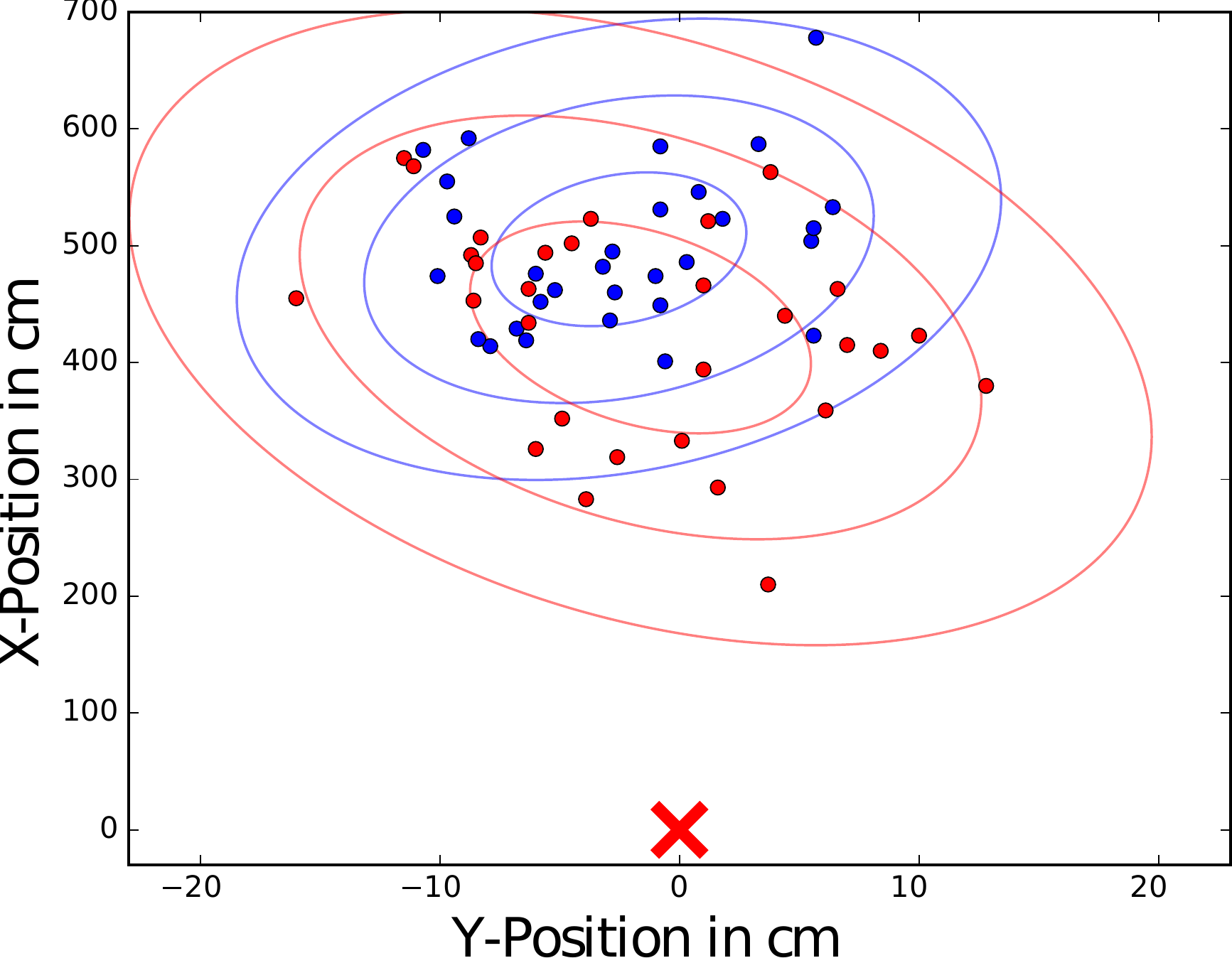}
    \label{fig:compare_imitated_kick:a}
    }
    \subfigure[]{
    \includegraphics[width=0.475\textwidth]{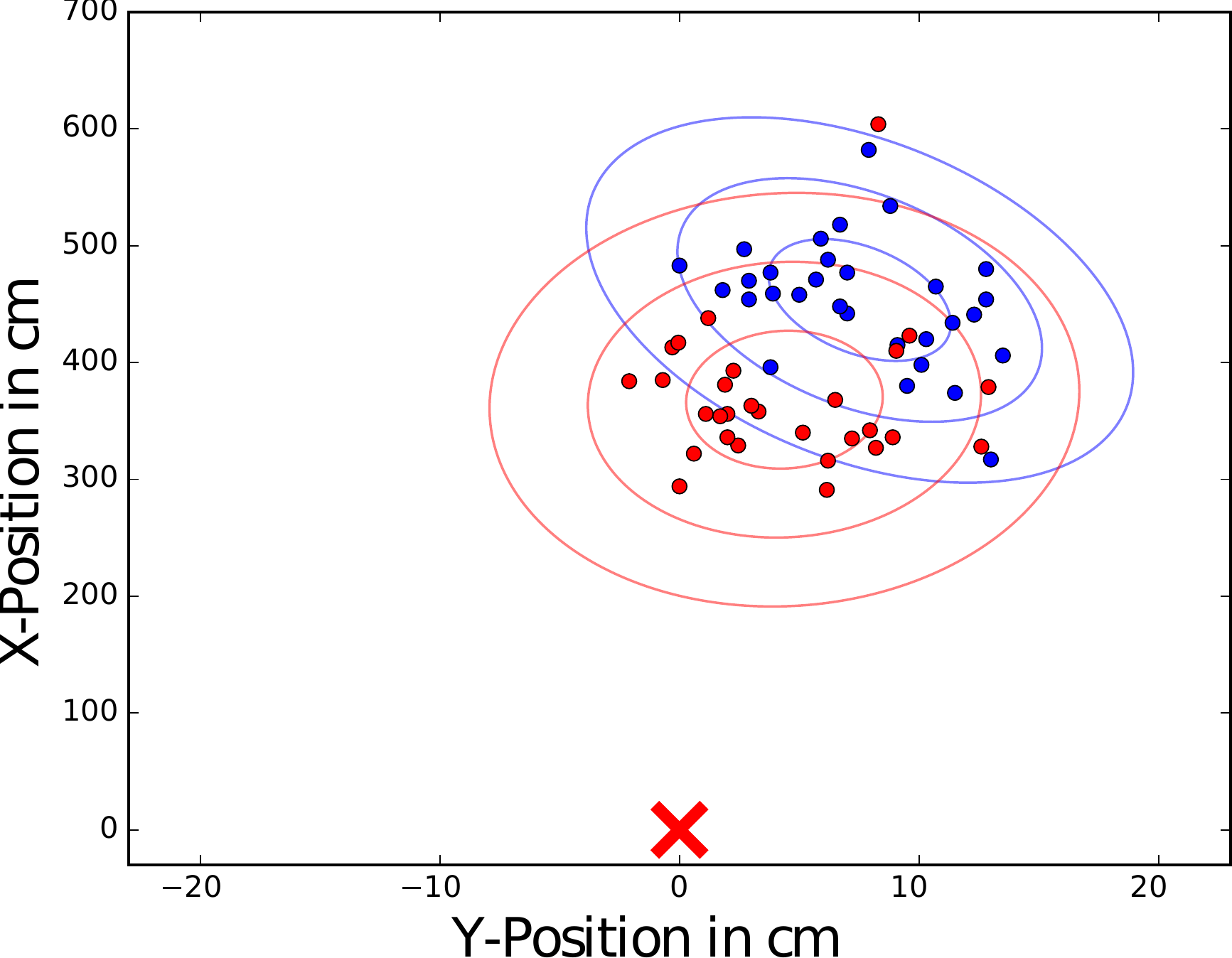}
    \label{fig:compare_imitated_kick:b}
    }

    \caption{Kick positions of the B-Human and imitated kick motions: The red cross is the kick origin. The dots are the positions where the balls came to a halt. The blue dots originate from the B-Human kick, the red dots from the imitated kick. (a) shows the results of kicks with the left foot while (b) shows the right foot.}\label{fig:compare_imitated_kick}
\end{figure}

To compare the performance of the kick motion to an existing one, we used imitation learning \cite{schaal2003control} to learn weights that imitate the kick motion of team B-Human of 2015\cite{codeRel2015,BIKE-RoboCup-2010} and executed 30 kicks with each leg and each kick motion. 
The results can be seen in \tabref{tab:imitatedKickResults} and \figref{fig:compare_imitated_kick}.

To test the generalization qualities of the kick motion, we conducted four experiments with different ball positions and velocities. In the first experiment, the ball is positioned 65 mm to the left. In the second experiment it is moved 80 mm forward, the third and fourth experiment reduced the kick velocity by $1/4$ and $1/2$ of the original kick velocity respectively. The results can be seen in \tabref{tab:generalizationResults}. To reach the position of the left ball, the robot had to fully stretch the leg. Therefore, the knee motor could not be used to generate a forward force, thereby significantly reducing the reached kick distance. The other experiments show a reasonable scaling towards the desired kick distance. Videos showing the kick generalization can be found at \url{https://youtu.be/g73pPCWcQvw} and \url{https://youtu.be/eANtiAiMmTg}.

\begin{table}[tb]

  \begin{center}
    \begin{tabular}{|l|l|l|l|l|}
    \hline
                        & Left ball  & Forward ball     & $3/4$ speed  & $1/2$ speed\\\hline\hline
Avg. distance           & 1.31 m          & 3.29 m           & 2.4 m        & 1.83 m\\
Avg. angular deviation  & $5.91^\circ$    & $4.98^\circ$     & $5.83^\circ$ & $4.73^\circ$\\
Avg. ball location (cm) & (12.63, 130.93) & (22.87, 328.0)   &(5.44, 239.16)&(2.96, 182.90)\\     
Royston H-value         &     6.17        & 1.15             & 3.65         &  3.43\\
Royston p-value         &    0.04         & 0.56             & 0.144        & 0.142\\
Is normal distributed   &    No           & Yes              & Yes          & Yes \\\hline
    \end{tabular}
  \end{center}
  \caption{Kick distance results for generalized kicks. The data of the left generalization contained one outlier. If it is removed, the result is normally distributed.}
  \label{tab:generalizationResults}
\end{table}

\section{Conclusion}
\label{sec:conclusion}
We presented a kick motion for the NAO robot that can imitate arbitrary kick trajectories and adapt them to different ball positions as well as different kick velocities. The kick motion is modeled using a slightly modified variant of DMPs. While executing the kick, the robot is kept dynamically stable using a ZMP preview controller. Additionally, we proposed a model of the NAO's motors and used it to improve the calculation of the ZMP.

 We have shown that this method of generating kick motions works but cannot kick the ball as far as a manually tuned motions. However, they are more versatile. The ball does not need to be placed perfectly to be kicked and the kick speed can be adjusted. Additionally, the underlying DMPs are easy to extend and lend themselves well to a multitude of optimization algorithms \cite{kober2013reinforcement}.
 
The kick motion presented in this paper has been successfully used in the corner kick challenge competition at RoboCup 2015.

\subsection*{Acknowledgement}
We would like to thank the members of the team B-Human for providing the
software framework for this work. 
\bibliography{bibliography}

\end{document}